\documentclass{article}
\usepackage[utf8]{inputenc}
\usepackage[dvipsnames]{xcolor}
\usepackage{graphicx}
\usepackage{amsmath}
\usepackage{bbm}
\usepackage{booktabs}
\usepackage{amsfonts}
\usepackage{selectp}
\usepackage{microtype}
\usepackage{graphicx}
\usepackage{subfigure}
\usepackage{booktabs} 

\RequirePackage{natbib}

\usepackage[margin=1in]{geometry}
\usepackage{subfiles}

\usepackage[colorlinks,citecolor=blue,urlcolor=blue,linkcolor=blue,linktocpage=true]{hyperref}

\usepackage[T1]{fontenc}

\definecolor{chng}{rgb}{0.0, 0.0, 0.0}
\newcommand{\change}[1]{\textcolor{chng}{#1}}
\usepackage{multirow}
\usepackage[colorinlistoftodos,textwidth=3.0cm,textsize=tiny]{todonotes}

\title{Disentangling the Mechanisms Behind\\ Implicit Regularization in SGD}

\author{Zachary Novack\footnote{\texttt{znovack@ucsd.com}, UC San Diego.},\; 
        Simran Kaur\footnote{\texttt{skaur@princeton.edu}, Princeton University.},\; 
        Tanya Marwah\footnote{\texttt{tmarwah@andrew.cmu.edu}, Carnegie Mellon University},\;
        Saurabh Garg\footnote{\texttt{sgarg2@andrew.cmu.edu}, Carnegie Mellon University},\; 
        Zachary C. Lipton \footnote{\texttt{zlipton@andrew.cmu.edu}, Carnegie Mellon University}}
\date{}

\begin{document}
\maketitle
\newcommand{\btheta}{\boldsymbol{\theta}}
\newcommand{\bdelta}{\boldsymbol{\delta}}
\newcommand{\bx}{\mathbf{x}}
\newcommand{\bz}{\mathbf{z}}
\newcommand{\calL}{\mathcal{L}}
\newcommand{\calB}{\mathcal{B}}
\newcommand\norm[1]{\left\lVert#1\right\rVert}
\newcommand{\amlgn}{Average Micro-batch Gradient Norm }
\definecolor{red}{rgb}{0.0, 0.0, 0.0}

\begin{abstract}
    A number of competing hypotheses have been proposed to explain 
    \emph{why} small-batch Stochastic Gradient Descent (SGD)
    leads to improved generalization over the full-batch regime, 
    with recent work crediting the implicit regularization 
    of various quantities throughout training.
    However, to date, empirical evidence assessing
    the explanatory power of these hypotheses is lacking.
    In this paper, we conduct an extensive empirical
    evaluation, focusing on the ability 
    of various theorized mechanisms to close 
    the small-to-large batch generalization gap.
    Additionally, we characterize how the quantities 
    that SGD has been claimed to (implicitly) regularize
    change over the course of training.
    By using \emph{micro-batches}, i.e.
    disjoint smaller subsets of each mini-batch, 
    we empirically show that explicitly penalizing 
    the gradient norm or the Fisher Information Matrix trace, 
    averaged over micro-batches, in the large-batch regime 
    recovers small-batch SGD generalization, 
    whereas Jacobian-based regularizations fail to do so.  
    This generalization performance is shown
    to often be correlated with how well the 
    regularized model's gradient norms 
    resemble those of small-batch SGD. 
    We additionally show that this behavior breaks down 
    as the micro-batch size approaches the batch size. 
    Finally, we note that in this line of inquiry,
    positive experimental findings on CIFAR10 
    are often reversed on other datasets like CIFAR100, 
    highlighting the need to test hypotheses 
    on a wider collection of datasets.
\end{abstract}

\section{Introduction}

While small-batch SGD has frequently 
been observed to outperform large-batch SGD
\citep{DBLP:journals/corr/abs-2109-14119, DBLP:journals/corr/KeskarMNST16, DBLP:journals/corr/abs-1804-07612, DBLP:journals/corr/abs-2101-12176, DBLP:journals/corr/abs-1906-07405, DBLP:journals/corr/abs-1711-04623, DBLP:conf/nips/WuME18,wen2020empirical,noisehanc},
the upstream cause for this generalization gap 
is a contested topic, approached 
from a variety of analytical perspectives \citep{DBLP:journals/corr/GoyalDGNWKTJH17, DBLP:journals/corr/abs-1906-07405,DBLP:journals/corr/abs-2109-14119,lee2022implicit}. 
Initial work in this field has generally 
focused on the learning rate to batch-size ratio
\citep{DBLP:journals/corr/KeskarMNST16, DBLP:journals/corr/abs-1804-07612,DBLP:journals/corr/GoyalDGNWKTJH17,lrbs1,lrbs2,lrbs3} 
or on recreating stochastic noise via mini-batching 
\citep{DBLP:journals/corr/abs-1906-07405, DBLP:journals/corr/abs-1711-04623,zhu2019anisotropic,noisehanc,cheng2020stochastic,simsekli2019tailindex,posneg},
whereas
recent works
have pivoted 
focus on understanding
how mini-batch SGD may \emph{implicitly regularize} 
certain quantities that improve generalization
\citep{DBLP:journals/corr/abs-2109-14119, DBLP:journals/corr/abs-2009-11162, DBLP:journals/corr/abs-2101-12176, lee2022implicit, DBLP:journals/corr/abs-2012-14193}.

\begin{figure}[ht]
    \centering\includegraphics[width=\textwidth]{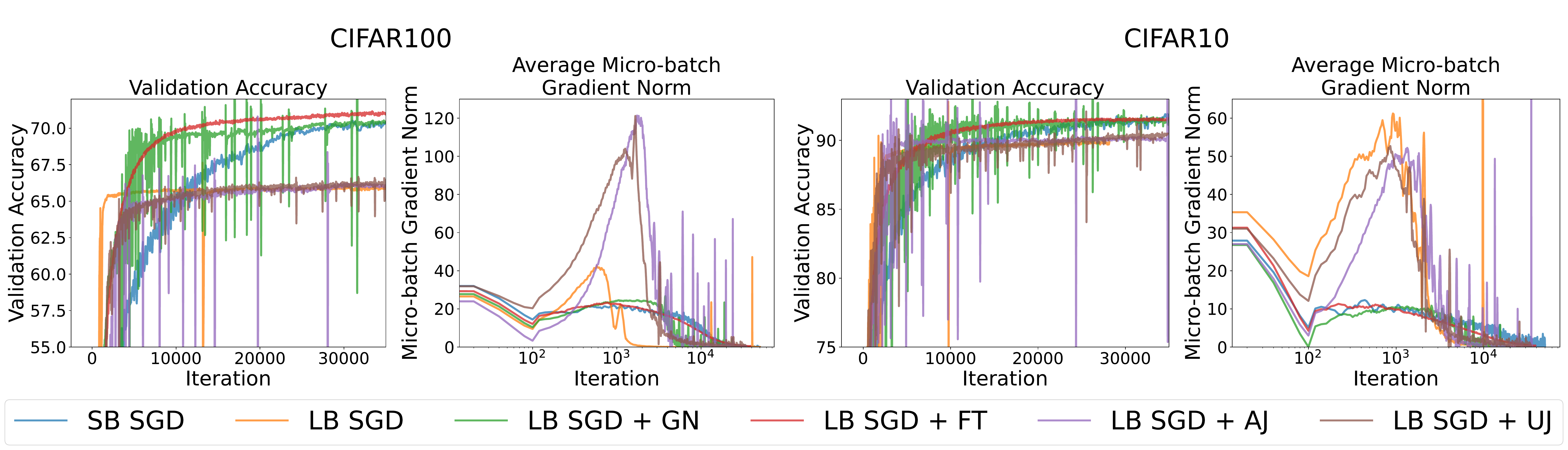}
    \caption{Validation Accuracy and Average Micro-batch ($|M| = 128$) Gradient Norm for CIFAR10/100 Regularization Experiments, averaged across runs (plots also smoothed for clarity). In both datasets, Gradient Norm (GN) and Fisher Trace (FT) Regularization mimic the average micro-batch gradient norm behavior of SGD during early training and effectively recover generalization performance \textcolor{red}{(within a small margin of error)}, whereas both Average and Unit Jacobian (AJ and UJ) fail to do so.}
    \label{fig:tmp-expl-reg}
\end{figure}

In this paper, we provide a careful empirical analysis 
of how these competing regularization theories 
compare to each other
as assessed by how well the prescribed interventions,
when applied in the large batch setting,
recover SGD's performance. 
Additionally, we study their similarities and differences 
by analyzing the evolution of the regularized quantities
over the course of training.

Our main contributions are the following:
\begin{enumerate}
    \item \textcolor{red}{By utilizing micro-batches (i.e. disjoint subsets of each mini-batch)}, we find that explicitly regularizing
    either the average micro-batch gradient norm 
    \citep{DBLP:journals/corr/abs-2109-14119, DBLP:journals/corr/abs-2009-11162}
    or Fisher Information Matrix trace 
    \citep{DBLP:journals/corr/abs-2012-14193}
    (equivalent to the average gradient norm when labels
    are drawn from the predictive distribution, 
    detailed in Section \ref{sec:fisheq})
    in the large-batch regime 
    fully recovers small-batch SGD generalization performance,
    but using Jacobian-based regularization \citep{lee2022implicit} 
    fails to recover small-batch SGD performance
    (see Figure \ref{fig:tmp-expl-reg}).
\item We show that the generalization performance 
    is strongly correlated with how well 
    the trajectory of the average micro-batch gradient norm during training 
    \emph{mimics} that of small-batch SGD,
    but that this condition is not necessary
    for recovering performance in some scenarios.
    The poor performance of Jacobian regularization,
    which enforces either uniform or fully random weighting 
    on each class and example (see Section \ref{sec:Jaceq}), 
    highlights that the beneficial aspects
    of average micro-batch gradient norm 
    or Fisher trace regularization 
    may come from the loss gradient's ability 
    to adaptively weight outputs 
    on the per example and per class basis. 
\item We demonstrate 
    that the \change{generalization} benefits of both successful methods
    \change{no longer hold when} the micro-batch size \change{is closer to} the actual batch size. 
    \change{We too} subsequently \change{show that in this regime} the average 
    micro-batch gradient norm behavior \change{of both previously successful methods}
    differs significantly from the small-batch SGD case.
\item We highlight a high-level issue 
    in modern empirical deep learning research: 
    Experimental results that hold on CIFAR10 
    do not necessarily carry over to other datasets.
    In particular, we focus on a technique
    called \emph{gradient grafting} \citep{DBLP:journals/corr/abs-2002-11803}, 
    \textcolor{red}{which has been shown to improve generalization 
    for adaptive gradient methods.
    By looking at its behavior for normal
    SGD and GD, }
    we show that gradient grafting
    recovers small-batch SGD generalization's performance on CIFAR10 but fails in CIFAR100,
    arguing that research in this line should prioritize
    experiments on a larger and diverse range of benchmark datasets.
\end{enumerate}

\section{Prior Work and Preliminaries}\label{sec:prior}
In neural network training,
the choice of batch size 
(and learning rate)
heavily influence generalization.
In particular, researchers have found that
opting for small batch sizes 
(and large learning rates)
improve a network's ability to generalize 
\citep{DBLP:journals/corr/KeskarMNST16, DBLP:journals/corr/abs-1804-07612,DBLP:journals/corr/GoyalDGNWKTJH17,lrbs1,lrbs2,lrbs3} .
Yet, explanations for this phenomenon have long been debated.
While some researchers have
attributed the success of small-batch SGD
to gradient noise introduced by stochasticity and mini-batching \citep{DBLP:journals/corr/abs-1906-07405, DBLP:journals/corr/abs-1711-04623,zhu2019anisotropic,noisehanc,cheng2020stochastic,simsekli2019tailindex,posneg},
others 
posit that small-batch SGD
finds ``flat minima''
with low non-uniformity, 
which in turn boosts generalization
\citep{DBLP:journals/corr/KeskarMNST16, DBLP:conf/nips/WuME18,simsekli2019tailindex}.
Meanwhile, 
some works
credit 
the implicit regularization of 
quantities such as loss gradient norm, the Jacobian norm (i.e., the network output-to-weights gradient norm), and the Fisher Information Matrix
\citep{DBLP:journals/corr/abs-2109-14119, DBLP:journals/corr/abs-2009-11162, DBLP:journals/corr/abs-2101-12176, lee2022implicit, DBLP:journals/corr/abs-2012-14193}.

Recent works have shown that one can recover SGD generalization performance 
by training 
on 
a modified loss function that 
regularizes large loss gradients
\citep{DBLP:journals/corr/abs-2109-14119, DBLP:journals/corr/abs-2009-11162, DBLP:journals/corr/abs-2101-12176}.
While \citet{DBLP:journals/corr/abs-2101-12176} and \citet{ DBLP:journals/corr/abs-2009-11162}
expect
that training on this modified loss function 
with large micro-batch sizes
will be unable to recover SGD generalization performance,
we empirically verify this is the case. 

\textcolor{red}{To our knowledge,
we are the first to introduce the ``micro-batch``
terminology to denote component disjoint sub-batches
used in accumulated mini-batch SGD.
This choice was made to avoid overloading
the term ``mini-batch`` 
and thus clarify the work
done by \citet{DBLP:journals/corr/abs-2101-12176} and \citet{DBLP:journals/corr/abs-2009-11162}.
Note that here and for the rest of this paper,
we use \emph{Large-Batch} SGD as an approximation
for full-batch GD
due to the computational constraints of using
the full training set on each update.}
\change{We emphasize that throughout this paper, 
micro-batches are not meant as any sort of ``alternative"
to mini-batches, as they are purely an implementation feature
of gradient accumulation-based large-batch SGD.}
We additionally leverage the work done by \citet{DBLP:journals/corr/abs-2002-11803}, 
who propose the idea of \emph{grafting} as a meta-optimization algorithm, 
though in the paper their focus mostly rests on grafting adaptive optimization algorithms together, 
not plain mini-batch SGD.

\change{As a whole, our paper is situated as
a comparative analysis of multiple proposed regularization
mechanisms \citep{DBLP:journals/corr/abs-2109-14119, DBLP:journals/corr/abs-2009-11162, DBLP:journals/corr/abs-2101-12176, lee2022implicit, DBLP:journals/corr/abs-2012-14193} in a side-by-side empirical context, with additional
ablations over how minor design choices may affect the efficacy
of these proposed methods to close the generalization gap.}
We now discuss various implicit and 
explicit regularization mechanisms in more depth.
 
 \textbf{Setup and Notation \enspace}
 We primarily consider the case of a softmax classifier $f: \mathbb{R}^d \rightarrow \mathbb{R}^C$ (where $C$ is the number of classes) parameterized by some deep neural network with parameters $\btheta$. We let $\ell(\bx, y; \btheta)$ denote the standard cross entropy loss for example $\bx$ and label $y$, and let $\calL_{\calB}(\btheta) = \frac{1}{|\calB|}\sum_{(\bx, y) \in \calB} \ell(\bx, y; \btheta)$ denote the average loss over some batch $\calB$.
 \textcolor{red}{Note that throughout
this paper, the terms ``batch"
and ``mini-batch" 
are used interchangeably
to refer to $\calB$.}

\subsection{Average Micro-batch Gradient Norm Regularization}
As proposed by \change{\citet{DBLP:journals/corr/abs-2101-12176}}, 
we attempt to understand the generalization behavior of 
mini-batch SGD by how it implicitly regularizes
the norm of the \textbf{micro-batch} gradient,
$\|\nabla\calL_{M}(\btheta)\|$ for some \textcolor{red}{micro-batch $M \subseteq \calB$}.
In large-batch SGD, we accomplish this 
through \emph{gradient accumulation} 
(i.e. accumulating the gradients of many small-batches
to generate the large-batch update), 
and thus can add an explicit regularizer 
(described in \citet{DBLP:journals/corr/abs-2109-14119}) 
that penalizes the \emph{average} micro-batch norm. 
Formally, for some large-batch $\calB$ 
and component disjoint micro-batches $M \subseteq \calB$, 
let $\nabla_{\btheta}\calL_{M}(\btheta) = \frac{1}{|M|}\sum_{(\bx, y) \in M}\nabla_{\btheta}\ell(\bx,y;\btheta)$.
The new loss function is:
\begin{equation}\label{eq:amlgnloss}
    \calL_{\calB}(\btheta) + \lambda \frac{|M|}{|\calB|}\sum_{M \in \calB} \|\nabla_{\btheta} \calL_{M}(\btheta)\|^2.
\end{equation}
While this quantity can be approximated through finite differences, 
it can also be
optimized directly by backpropagation
using modern deep learning packages,
as we do in this paper. 

\textcolor{red}{Note that by definition,
we can decompose the regularizer term
into the product of the 
Jacobian of the network and the
gradient of the loss with respect to network output. Formally, for some network $f$ with $p$ parameters,
if we let $\bz = f(\bx;\btheta) \in \mathbb{R}^C$ be
the model output for some input $\bx$ and denote its
corresponding label as $y$,
then:}
\begin{equation}\label{eq:gnjac}
\textcolor{red}{\nabla_{\btheta}\ell(\bx, y; \btheta) = (\nabla_{\btheta}\bz) (\nabla_{\bz}\ell(\bx, y; \btheta))}
\end{equation}

\textcolor{red}{Where $\nabla_{\btheta}\bz \in \mathbb{R}^{p \times C}$
is the Jacobian of the network
and the second term is the \emph{loss-output} gradient.
We explicitly show this equivalence for the comparison
made in section \ref{sec:Jaceq}.}

\subsection{Average Micro-batch Fisher Trace Regularization}\label{sec:fisheq}
One noticeable artifact of Equation \ref{eq:amlgnloss} is its implicit reliance on the true labels $y$
to calculate the regularizer penalty.
\citet{DBLP:journals/corr/abs-2012-14193} shows that we can derive a similar quantity in the mini-batch SGD setting by penalizing 
the trace of the \emph{Fisher Information Matrix} $\mathbf{F}$, which is given by $\text{Tr}(\mathbf{F}) = \mathbb{E}_{\bx \sim \mathcal{X}, \hat{y} \sim p_{\btheta}(y \mid \bx)} [\| \nabla_{\btheta} \ell(\bx, \hat{y}; \btheta)\|^2]$,
where $p_{\btheta}(y \mid \bx)$ is the
predictive distribution of the model
at the current iteration and 
$\mathcal{X}$ is the data distribution. 
We thus extend their work to the accumulated large-batch regime
and penalize an approximation of the \emph{average} Fisher trace 
over micro-batches: if we let $\widehat{\calL}_{M}(\btheta) = \frac{1}{|M|}\sum_{\bx \in M,  \hat{y} \sim p_{\btheta}(y \mid \bx)} \ell(\bx, \hat{y}; \btheta)$, 
then our penalized loss is
\begin{equation}\label{eq:fishloss}
    \calL_{\calB}(\btheta) + \lambda \frac{|M|}{|\calB|}\sum_{M \in \calB} \|\nabla_{\btheta} \widehat{\calL}_{M}(\btheta)\|^2.
\end{equation}
The only difference between the expressions in 
Equation \ref{eq:amlgnloss} and Equation \ref{eq:fishloss} is that
the latter now uses labels sampled from the \emph{predictive} distribution,
rather than the true labels, 
to calculate the regularizer term. 
As with Equation \ref{eq:amlgnloss}, we can directly backpropagate
using this term in our loss equation. 

\textcolor{red}{Like, in equation \ref{eq:gnjac}, we can decompose the regularizer term as:}

\begin{equation}\label{eq:ftjac}
    \textcolor{red}{\ell(\bx, \hat{y}; \btheta) = (\nabla_{\btheta} \bz)(\nabla_{\bz}\ell(\bx, \hat{y}; \btheta))}
\end{equation}

\textcolor{red}{Where the second term is another loss-output gradient.}

\textcolor{red}{\citet{DBLP:journals/corr/abs-2012-14193} observes that 
models with poor generalization
typically show a large spike
in the Fisher Trace during
the early phases of training},
which they coin as \emph{Catastrophic Fisher Explosion}. 
In Figure \ref{fig:tmp-expl-reg}, 
we show that this behavior also occurs
when looking at the average Micro-Batch gradient norm.

\subsection{Jacobian Regularization}\label{sec:Jaceq}

\textcolor{red}{Given the decompositions shown in equations
\ref{eq:gnjac} and \ref{eq:ftjac},}
it is unclear in either case whether the Jacobian term
is the sole source of possible generalization benefit,
or if the loss-output gradient is also needed.
To disentangle this effect, we borrow from \citet{lee2022implicit}
and use the \emph{average} and \emph{unit} Jacobian regularization losses: 
\begin{align}\label{eq:JacsAvg}  
    \calL_{\calB}(\btheta) + \lambda \frac{|M|}{|\calB|}\sum_{M \subseteq \calB} \|J_{\text{avg}}(M)\|^2, 
    \quad &\quad 
    J_{\text{avg}}(M)=\frac{1}{|M|}\sum_{(\bx, y) \in M} (\nabla_{\btheta} \bz)(\frac{1}{C}\mathbbm{1}),
\end{align}
\begin{align}\label{eq:JacsUnif}
    \calL_{\calB}(\btheta) + \lambda \frac{|M|}{|\calB|}\sum_{M \subseteq \calB} \|J_{\text{unit}}(M)\|^2,
    \quad &\quad 
    J_{\text{unit}}(M)=\frac{1}{|M|}\sum_{(\bx, y) \in M} (\nabla_{\btheta} \bz) (\mathbf{u}),
\end{align}
where $\mathbf{u} \in \mathbb{R}^C$ is randomly sampled
from the unit hypersphere (i.e. $\|\mathbf{u}\|_2 = 1$),
and $\mathbf{u}$ is sampled once per micro-batch.
In words, the \emph{average} Jacobian case penalizes the Jacobian
with equal weighting on every class and every example,
while the \emph{unit} Jacobian case penalizes the Jacobian
with different but \emph{random} weighting on each class and example. 
\textcolor{red}{Note that the unit Jacobian penalty is
an unbiased estimator of the Frobenius norm
of the Jacobian $\|\nabla_{\btheta} \bz \|_F^2$,
which is an upper bound on its spectral norm
$\|\nabla_{\btheta} \bz \|_2^2$ (see \citet{lee2022implicit} for a more detailed theoretical analysis).}
\section{Explicit Regularization And Gradient Norm Trajectory}\label{sec:expresults}

These aforementioned explicit regularization mechanisms
have previously been investigated in limited empirical settings. 
To the best of our knowledge, \citet{DBLP:journals/corr/abs-2012-14193}
is the only work that has directly compared
some of these regularization mechanisms, 
but only did so in the context of
improving \emph{small-batch} performance. 
Like our work,
\citet{DBLP:journals/corr/abs-2109-14119} is centered on
the small-to-large batch generalization gap, 
but they do not focus \emph{solely} on 
the explicit regularization they propose
and do not include any analysis of 
the micro-batch gradient norm behavior during training.
In this work, we investigate
(i) how these regularizers effect generalization 
for an array of benchmarks and 
(ii) how such performance may correlate with 
the \emph{evolution} of the micro-batch gradient norm during training.

\clearpage
\subsection{Experimental Setup}

We first focus our experiments on the case of using a ResNet-18 \citep{https://doi.org/10.48550/arxiv.1512.03385},
with standard initialization and batch normalization, 
on the CIFAR10, CIFAR100, \textcolor{red}{Tiny-ImageNet, and SVHN} image classification benchmarks \citep{Krizhevsky09learningmultiple, Le2015TinyIV, svhn}. 
Additional experiments on different architectures are detailed in Appendix \ref{appendix:add}. 
Besides our small-batch ($\calB = 128$) and large-batch ($\calB = 5120$) SGD baselines,
we also train the networks in the large-batch regime using
(i) average Micro-batch Gradient Norm Regularization (GN);
(ii) average Micro-batch Fisher Trace Regularization (FT); 
and (iii) \emph{average} and \emph{unit} 
Micro-batch Jacobian Regularizations (AJ and UJ).
Note that for all the regularized experiments, 
we use a component micro-batch size equal
to the small-batch size (i.e. 128). 
In order to compare the \emph{best possible} performance
within each experimental regime, 
we separately tune the optimal learning rate $\eta$
and optimal regularization parameter $\lambda$
independently for each regime.
Additional experimental details
can be found in Appendix \ref{appendix:setup}.

\begin{table}
    \centering\textcolor{red}{
    \caption{ResNet18 Test Performance for Regularizer Penalties. Values shown are average test accuracies across two to three different initializations per experiment, with corresponding confidence intervals.}\label{tab:resresults}
    \begin{tabular}{lllll}
    \toprule
    Experiment & CIFAR10 & CIFAR100 & Tiny-ImageNet & SVHN \\
    \midrule
    SB SGD & 92.33 ($\pm 0.10$) & 71.01 ($\pm 0.27$) & 39.64 ($\pm 0.18$) & 93.69 ($\pm 0.12$) \\
    LB SGD & 90.00 ($\pm 0.11$) & 66.45 ($\pm 0.29$) & 27.71 ($\pm 0.09$) & 90.37 ($\pm 0.33$)\\
    GN & 91.98 ($\pm 0.03$) & 70.22 ($\pm 0.27$) & 37.78 ($\pm 0.07$) & 92.77 ($\pm 0.01$) \\
    FT & 91.79 ($\pm 0.05$) & 71.19 ($\pm 0.16$) & 40.25 ($\pm 0.02$) & 93.72 ($\pm 0.16$) \\
    AJ & 90.41 ($\pm 0.01$) & 65.95 ($\pm 0.33$) & 22.86 ($\pm 0.95$) & 91.76 ($\pm 0.11$) \\
    UJ & 90.46 ($\pm 0.20$) & 66.41 ($\pm 0.06$) & 26.07 ($\pm 0.54$) & 92.08 ($\pm 0.01$)\\
    \bottomrule
    \end{tabular}}
\end{table}

\subsection{Results}

\textbf{(i) Average micro-batch Gradient Norm and average Fisher trace Regularization recover SGD generalization \enspace}
\textcolor{red}{For CIFAR100, Tiny-ImageNet, and SVHN
we find that
we can fully recover small-batch SGD generalization performance by
penalizing the average micro-batch Fisher trace
and nearly recover performance by penalizing
the average micro-batch gradient norm
(with an optimally tuned regularization parameter $\lambda$, see Figure \ref{fig:tmp-expl-reg} and Table \ref{tab:resresults}). In CIFAR10, 
neither penalizing the gradient norm nor the Fisher trace
\emph{completely} recovers small-batch SGD performance,
but rather come within $\approx0.3\%$ and $\approx0.4\%$ (respectively)
the small-batch SGD performance and significantly improves
over large-batch SGD.}

We additionally find that using the micro-batch gradient norm 
leads to slightly faster per-iteration convergence 
but less stable training 
(as noted by the tendency for the model to exhibit random drops in performance), 
while using the Fisher trace
leads to slightly slower per-iteration convergence
but much more stable training (see Figure \ref{fig:tmp-expl-reg}).
This behavior may be due to the Fisher trace's ability 
to more reliably mimic the small-batch SGD micro-batch gradient norm behavior \emph{throughout} training, 
whereas penalizing the gradient norm effectively curbs the initial explosion 
but collapses to much smaller norm values as training progresses.

\textbf{(ii) Average and Unit Jacobian regularizations do not recover SGD generalization \enspace}
Observe 
in Table \ref{tab:resresults}
that we are unable to match SGD generalization performance
with either Jacobian regularization.
\change{In Section \ref{sec:prior} we showed that each
regularization method can be viewed as penalizing the
norm of the Jacobian matrix-vector product with \emph{some} $C$-dimensional vector.
Crucially, both the gradient norm and Fisher trace
regularizers use some form of loss-output gradient,
which is data-dependent and has no 
constraint on the weighting of each class,
while both Jacobian regularizers use data-independent
and comparatively simpler vectors.}
\change{Given the noticeable difference in generalization performance between the regularization methods that weight the Jacobian with the loss-output gradient
and those that do not, we indicate that 
the loss-output gradient may be crucial to
either applying the beneficial regularization effect itself
and/or stabilizing the training procedure.}
\section{Shortcomings and Extensions of Gradient Norm Regularization}

\subsection{Generalization Failure at Large Micro-Batch Sizes}
In both successful regularization regimes, 
namely the average micro-batch gradient norm 
and average fisher trace regularizers, 
there is an implicit hyperparameter:
the size of the micro-batch
used to calculate the regularization term. 
Note that this hyperparameter is a practical artifact
of modern GPU memory limits, 
as efficiently calculating higher-order derivatives
for large batch sizes is not feasible in standard 
autodifferentiation packages.
Consequently, gradient accumulation
(and the use of the average micro-batch regularizer,
rather than taking the norm over the entire batch) 
must be used on most standard GPUs 
\textcolor{red}{(more detailed hadware specifications can
be found in appendix \ref{appendix:setup})}. 

This restriction, however, may actually be beneficial,
as \citet{DBLP:journals/corr/abs-2109-14119, DBLP:journals/corr/abs-2009-11162, DBLP:journals/corr/abs-2101-12176} have noted that
they expect the benefits of gradient norm regularizers to break down
when the micro-batch size becomes too large.
To test this hypothesis, we return to the 
ResNet-18 in CIFAR100 and Tiny-ImageNet settings and 
increase the micro-batch size to 
as large as we could reasonably fit on a single GPU at $|M| = 2560$
in both the gradient norm and Fisher trace experiments. 
Additionally, we run experiments using a VGG11 \citep{https://doi.org/10.48550/arxiv.1409.1556}
on CIFAR10,
interpolating the micro-batch size from the small to large regimes. 
In both settings, 
we separately tune the learning rate $\eta$ and regularization coefficient $\lambda$ 
in each experiment to find the best possible generalization performance
in the large micro-batch regimes.

\begin{table}[h]
    \centering\textcolor{red}{
    \caption{Test Performance for ResNet-18  with Increased Micro-Batch Size. Small-batch SGD performances: CIFAR100 = $71.01$, Tiny-ImageNet = $39.64$.}\label{tab:largeresults}
    \begin{tabular}{llllll}
    \toprule
    Dataset & GN ($|M|=128$) & FT ($|M|=128$) & GN ($|M|=2560$) & FT ($|M|=2560$) \\
    \midrule
    CIFAR100 & 70.22 ($\pm 0.27$) & 71.19 ($\pm 0.16$) & 64.23 ($\pm 0.49$)& 65.44 ($\pm 0.76$)\\
    Tiny-ImageNet &37.78 ($\pm 0.07$) & 40.25 ($\pm 0.02$) & 31.96 ($\pm 0.56$) & 37.71 ($\pm 0.31$)\\
    \bottomrule
    \end{tabular}}
\end{table}

\begin{figure}[h]
    \centering\includegraphics[width=0.8\linewidth]{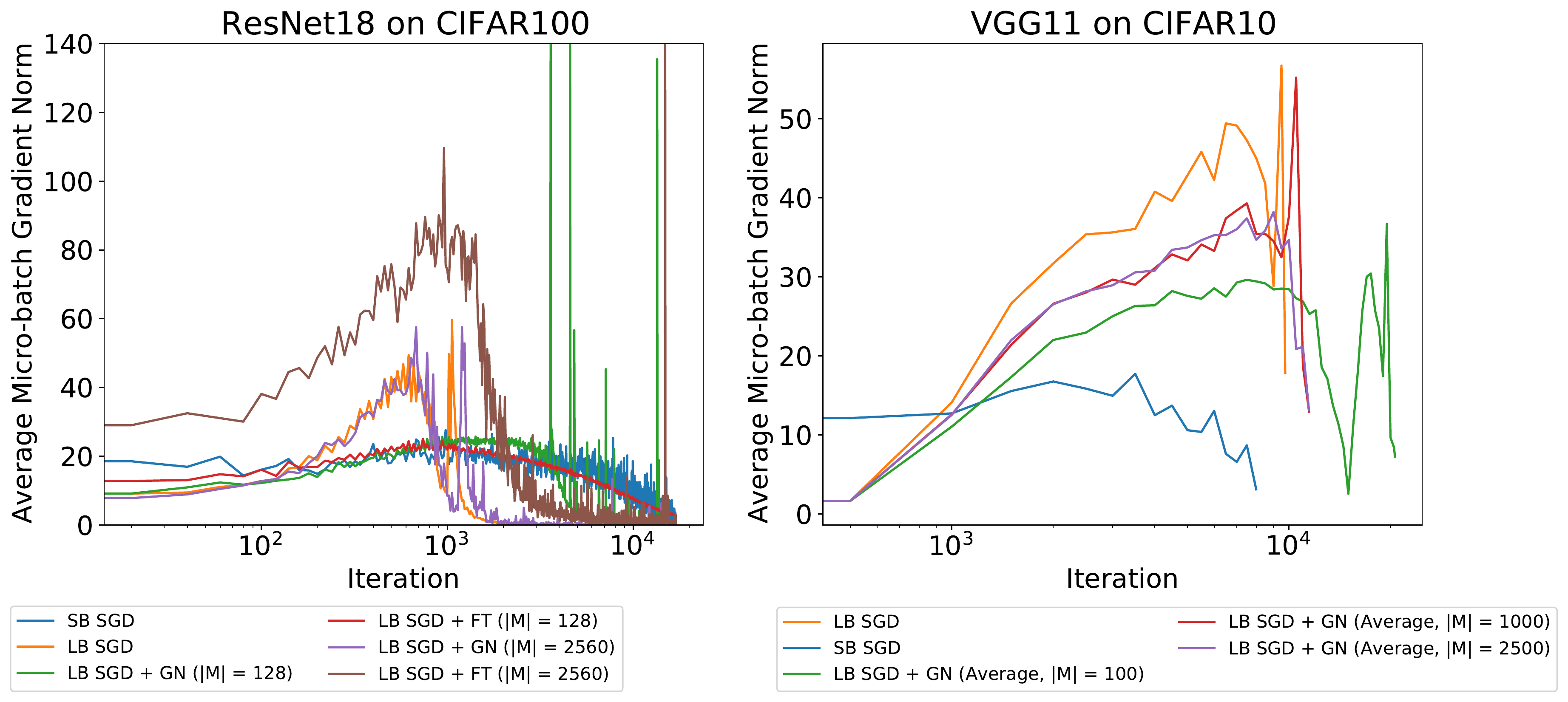}
    \caption{Average Micro-batch Gradient Norm for varying micro-batch sizes. In all experimental regimes, increasing the micro-batch size leads to a worse reconstruction of the SGD average micro-batch gradient norm behavior, especially in early training.}
    \label{fig:lb_failure}
\end{figure}

\begin{table}[h]
    \centering
    \caption{Test Performance for VGG11 (no batch-normalization) in CIFAR10 with Increased Micro-Batch Size}\label{tab:largeresults2}
    \begin{tabular}{lllllll}
    \toprule
    SB SGD & LB SGD & GN ($|M|=100$) & GN ($|M|=1000$) & GN ($|M|=2500$)\\
    \midrule
    78.19 & 73.90 & 76.89 ($\pm 0.72$) & 75.19 ($\pm 0.10$) & 75.11 ($\pm 0.29$) \\
    \bottomrule
    \end{tabular}
\end{table}

\textbf{Results \enspace}
We successfully show that such hypotheses mentioned in
\citet{DBLP:journals/corr/abs-2109-14119, DBLP:journals/corr/abs-2101-12176,DBLP:journals/corr/abs-2009-11162} hold true:
as the micro-batch size approaches the mini-batch size, 
\textcolor{red}{both regularization mechanisms lose the ability to
recover small-batch SGD performance}
(see Tables \ref{tab:largeresults} and \ref{tab:largeresults2}). Additionally, we note that using large micro-batch sizes
no longer effectively mimics the average micro-batch gradient norm
behavior of small-batch SGD, thus supporting our claim that
matching this quantity throughout training is of key 
importance to recovering generalization performance (Figure \ref{fig:lb_failure}).

\subsection{Sample Micro-batch Gradient Norm Regularization}\label{sec:samp}
One potential practical drawback of these gradient-based regularization terms is the relatively high computation cost needed to calculate the second-order gradients for every component micro-batch. Instead of penalizing the average micro-batch gradient norm, we can penalize \textit{one} micro-batch gradient norm. For some large batch $\calB$ and fixed sample micro-batch $S$ from batch $\calB$, we define the modified loss function
\begin{equation}
   \calL_{\calB}(\btheta) + \lambda \|\nabla_{\btheta} \calL_{S}(\btheta)\|^2.
\end{equation}

\begin{figure}[h]
    \centering
    \includegraphics[width=\textwidth]{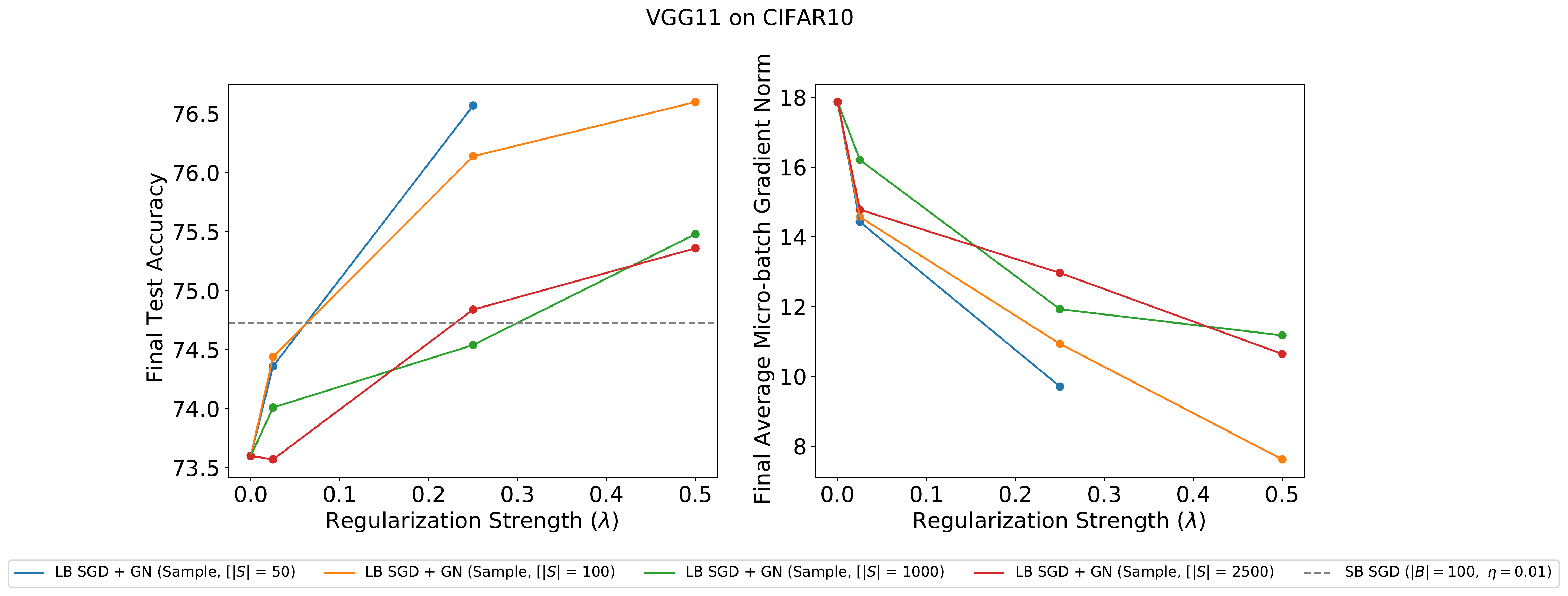}
    \caption{Explicitly regularizing the sample loss gradient norm recovers SGD test accuracy}
    \label{fig:sgnp-5000-0.01}
\end{figure}

\textbf{Results \enspace}
In Figure \ref{fig:sgnp-5000-0.01}, we plot the final test accuracy (left column) and the average gradient norm (right column) as a function of $\lambda$.
We observe that both 
a larger $\lambda$ 
and a smaller micro-batch size $|S|$
boost test accuracy.
Furthermore, we find that with the ``optimal'' $\lambda$ and micro-batch size $|S|$, the final test accuracy for sample micro-batch gradient norm regularization is close to (and sometimes better) than the final test accuracy for SGD.
Just as we observed with the \emph{average} Micro-batch Gradient Norm regularization,
generalization benefits diminish as the sample micro-batch size approaches the mini-batch size.

\section{Is mimicking SGD Gradient Norm Behavior necessary for generalization?}

As seen in Figure \ref{fig:tmp-expl-reg},
the trajectory of the average micro-batch gradient norm 
during training, 
and its similarity to that of small-batch SGD
especially in the early stages of training,
is strongly correlated with generalization performance.
Furthermore, we have observed that
models with \emph{poor} generalization performance
tend to exhibit the characteristic ``explosion''
during the early phase of training
and quickly plummet to
average micro-batch gradient norm values
much smaller than seen in small-batch SGD.
That being said, it is not immediately clear whether
recreating the micro-batch norm trajectory
of small-batch SGD 
is \emph{necessary} for ensuring good
generalization performance 
\change{(i.e. whether good generalization directly
implies gradient norm behavior similar to SGD)}.

To test this hypothesis, 
we empirically validate
an orthogonal vein of optimization methods
\change{that do not explicitly regularize
the micro-batch gradient norm during training}
for their ability to close the small-to-large batch generalization gap, 
and whether they too mimic the average micro-batch norm
trajectory of small-batch SGD.

\subsection{External and Iterative Grafting and Normalized Gradient Descent}

Inspired by the work of \citet{DBLP:journals/corr/abs-2002-11803}, 
we proposed to use \emph{gradient grafting} 
in order to control the loss gradient norm
behavior during training. 
Formally, for any two different optimization algorithms 
$\mathcal{M}, \mathcal{D}$, 
the grafted updated rule is arbitrarily:
\begin{align}\label{eq:grafting}
    g_\mathcal{M} &= \mathcal{M}(\btheta_k), \quad \quad g_\mathcal{D} = \mathcal{D}(\btheta_k) \nonumber\\
    \btheta_{k+1} &= \btheta_k - \|g_\mathcal{M}\|\frac{g_\mathcal{D}}{\|g_\mathcal{D}\|}
\end{align}
In this sense, 
$\mathcal{M}$ controls the \emph{magnitude} of the update step
and $\mathcal{D}$ controls the \emph{direction}. 
We first propose \textbf{Iterative Grafting}, 
wherein $\mathcal{M}(\btheta_k) = \eta \nabla \calL_M(\btheta_k)$ and $\mathcal{D}(\btheta_k) = \nabla \calL_{\mathcal{B}}(\btheta_k)$, 
where $M \in \calB$ is sampled uniformly
from the component micro-batches at every update.
In words, at every update step we take the large batch gradient,
normalize it, 
and then rescale the gradient by the norm
of one of the component micro-batch gradients.

Additionally, 
we propose \textbf{External Grafting}, where $\mathcal{M}(\btheta_k) = \eta \nabla \calL_M(\btheta_{k'})$ and $\mathcal{D}(\btheta_k) = \nabla \calL_{\mathcal{B}}(\btheta_k)$.
Here, we use $\nabla \calL_{B}(\btheta_{k'})$ to denote
the gradient norm at step $k$ 
from a \emph{separate small-batch SGD training run}. 
\textcolor{red}{We propose this experiment to make a comparison
with the Iterative Grafting case,
since here the implicit step length schedule
is independent of the current run,
while with Iterative grafting the schedule
depends upon the current training dynamics.}

Aside from grafting algorithms,
which define the implicit step length schedule at every step,
we also consider the situation where
the step length is fixed throughout training through \textbf{normalized gradient descent (NGD)}
\citep{https://doi.org/10.48550/arxiv.1507.02030}, wherein $\mathcal{M}(\btheta_k) = \eta$ and $\mathcal{D}(\btheta_k) = \eta \nabla \calL_{\mathcal{B}}(\btheta_k)$.

\begin{table}
    \centering
    \caption{Test Performance for Grafting / NGD Experiments}\label{tab:graftresults}
    \begin{tabular}{lllllll}
    \toprule
    Dataset & Model & SB SGD & LB SGD & EG & IG & NGD \\
    \midrule
    \multirow{2}{4em}{CIFAR10} & ResNet18 & 92.33 & 89.99 & 92.12 & 92.16 & 92.10\\
    & VGG16 w/Batch-Norm & 89.56 & 86.97 & 88.65 & 89.06 & 89.39\\
    \multirow{2}{5em}{CIFAR100} & ResNet18 & 71.21 & 66.17 & 68.3 & 68.4 & 66.83 \\
    & VGG16 w/Batch-Norm & 64.26 & 55.94 & 59.71 & 63.48 &  58.05\\
    \bottomrule
    \end{tabular}
\end{table}

\begin{figure}[h]
    \centering
    \includegraphics[width=0.8\textwidth]{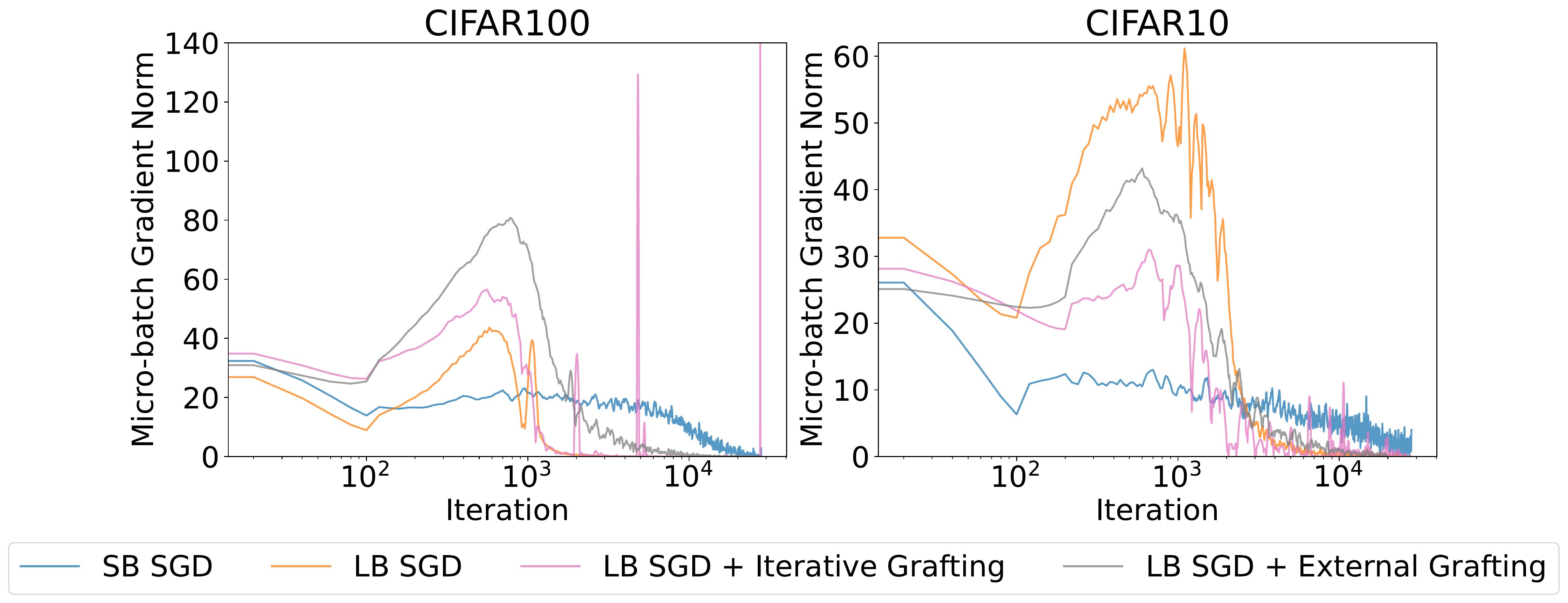}
    \caption{Average Micro-batch Gradient Norm in Grafting Experiments for ResNet-18 (Plots smoothed for clarity). In both scenarios, irrespective of generalization performance, the grafting experiments do not mimic the SGD average micro-batch gradient norm behavior. }\label{fig:graftplots}
\end{figure}

\textbf{Results \enspace}
We find that both forms of grafting and NGD can recover the generalization performance of SGD \emph{in some model-dataset combinations} (see Table \ref{tab:graftresults}). Namely, though grafting / NGD seems to work quite well in CIFAR10, no amount of hyperparameter tuning was able to recover the SGD performance for either model in CIFAR100. That being said, in the CIFAR10 case we see (in Figure \ref{fig:graftplots}) that the grafting experiments (and NGD, not pictured) \emph{do not} replicate the same average mini-batch gradient norm behavior of small-batch SGD despite sometimes replicating its performance. This thus gives us solid empirical evidence that while controlling the average mini-batch gradient norm behavior through explicit regularization may aide generalization, it is not the only mechanism in which large-batch SGD can recover performance.

\subsection{Wider Implications}\label{sec:imps}
The stark disparity in performance
between the CIFAR10 and CIFAR100 benchmarks 
are of key importance. 
These differences may be explained
by the much larger disparity between 
the mid-stage average micro-batch gradient norm behavior
in the CIFAR100 case than in the CIFAR10 case (see Figure \ref{fig:graftplots}). This situation highlights a possible cultural issue
within the deep learning community: 
there is a concerning trend of papers in the deep learning field
that cite desired performance on \emph{CIFAR10},
and no harder datasets,
as empirical justification for any posed theoretical results \citep{https://doi.org/10.48550/arxiv.2202.02831, DBLP:journals/corr/abs-2109-14119, DBLP:journals/corr/abs-2002-11803, DBLP:journals/corr/abs-2101-12176, DBLP:journals/corr/abs-2009-11162, cheng2020stochastic}. 
Given the continued advancement of state-of-the-art deep learning models,
we argue that it is imperative that baselines
like CIFAR100 and ImageNet
are adopted as the main standard for empirical verification,
so that possibly non-generalizable results
(as the grafting / NGD results would have been had we stopped at CIFAR10)
do not fall through the cracks in the larger community \textcolor{red}{(see Appendix \ref{appendix:pgd} for more information)}.
\section{Discussion \& Conclusion}

In this paper, we provide a holistic account
of how the proposed regularization mechanisms
\citep{DBLP:journals/corr/abs-2109-14119, DBLP:journals/corr/abs-2009-11162, DBLP:journals/corr/abs-2101-12176, lee2022implicit, DBLP:journals/corr/abs-2012-14193}
compare to each other in performance and gradient norm trajectory, 
and additionally show the limitations of
this analytical paradigm for explaining
the root cause of generalization.
%
Our results with regards to the relative poor performance
of the Jacobian-based regularizations
somewhat conflict with the results of \citet{lee2022implicit}, 
which shows positive results on using the unit Jacobian regularization
with respect to improving performance \emph{within the same batch-size regime}.
We attribute this difference to the fact that 
\citet{lee2022implicit}
is not
concerned with cases where the small-to-large batch generalization gap exists,
which is our main focus. 

In light of this prior work,
more research should be done to disentangle
the exact effect that 
implicitly regularizing the loss-output gradient
has on generalization performance.
Next,
given the success of 
average micro-batch gradient norm 
and average micro-batch Fisher trace regularization 
(especially with small micro-batches), 
future work should 
leverage these regularization mechanisms
to investigate
the possibility of
ameliorating generalization,
while
improving time efficiency,  
by 
taking advantage of 
high resource, parallelizable settings.
%
\textcolor{red}{We also show that
experimental findings on CIFAR10 
may no longer hold in CIFAR100,
which sheds light on 
a wider implication
for the research community.
Namely,
we urge researchers to adapt the
practice of evaluating 
empirical hypotheses on
a more widespread, complex set of benchmarks.}


We acknowledge that performance
in each experiment could possibly
be improved by progressively finer hyperparameter
tuning, \textcolor{red}{though we are confident that our core results
would continue to hold in such situations
given the extensive hyperparameter searches
performed for each experiment.
As a whole, the present research
helps to shed light on the mechanisms
behind SGD's generalization properties
through implicit regularization,
and offers robust fixes to the
generalization issue at high batch-sizes.}


\bibliography{references}

\clearpage
\appendix
\section{Appendix}

\subsection{Additional Regularization Experiments}\label{appendix:add}

Aside from the main results using a ResNet-18, we additionally ran the regularization experiments with a VGG11 \citep{https://doi.org/10.48550/arxiv.1409.1556} without batch normalization on CIFAR10. Results are shown below:

\begin{table}[h]
    \centering
    \caption{VGG11 (no batch-normalization) Test Performance for Regularizer Penalties}\label{tab:resresults2}
    \begin{tabular}{lllllll}
    \toprule
    Dataset & SB SGD & LB SGD & GN & FT & AJ &  UJ \\
    \midrule
    CIFAR10 & 78.19 & 73.90 & 77.62 & 79.10 & 74.09 & N/A \\
    \bottomrule
    \end{tabular}
\end{table}

Consistent with our earlier observations (see Section \ref{sec:expresults}), 
we find that average micro-batch gradient norm and average Fisher trace regularization nearly recover SGD generalization performance,
whereas average Jacobian regularization does not.

\subsection{Sample Micro-batch Gradient Norm Regularization (Continued)}\label{appendix:samp}

\begin{figure}[h]
    \centering
    \includegraphics[width=\textwidth]{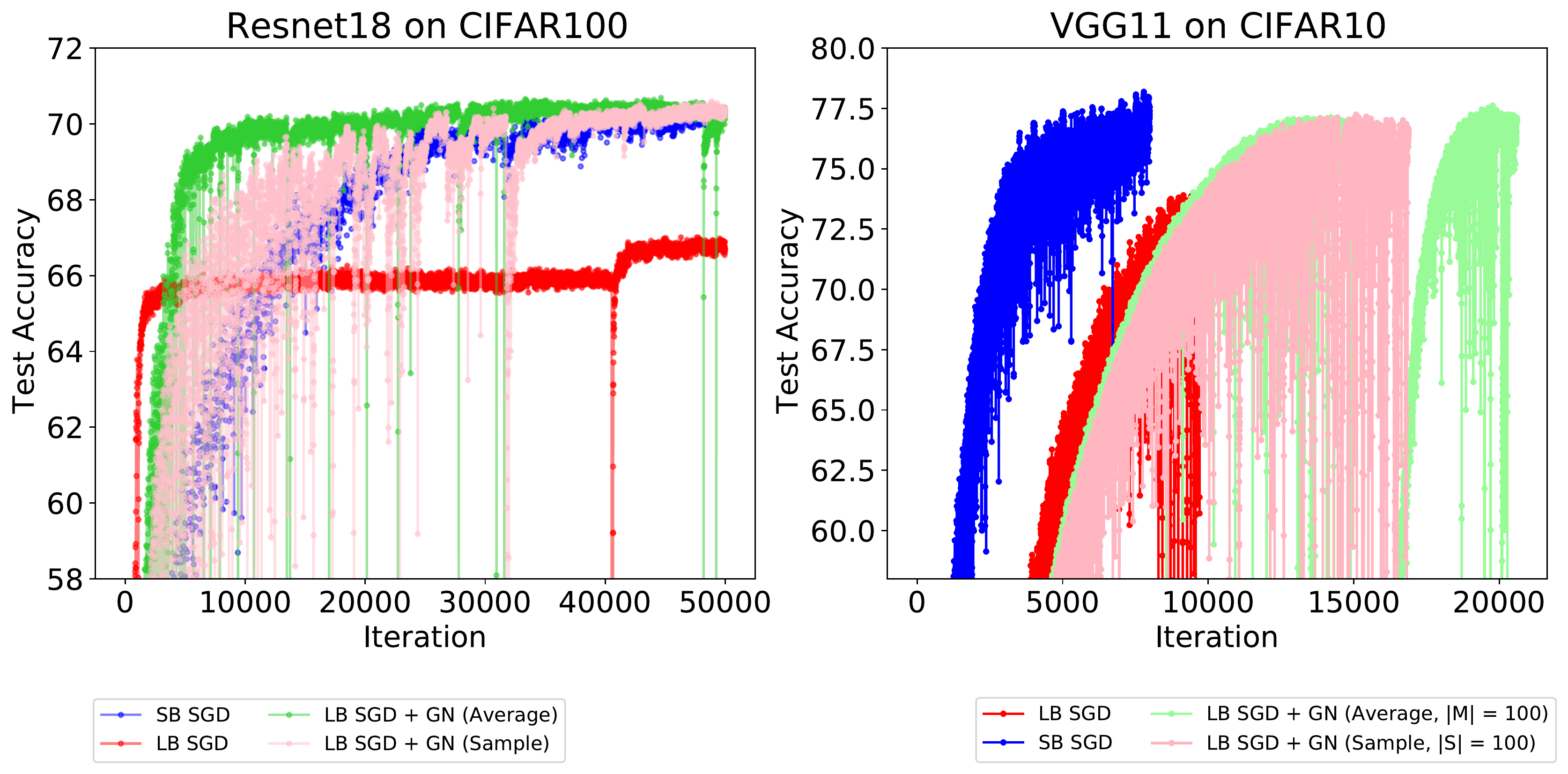}
    \caption{Explicitly regularizing the average loss gradient norm over a sample recovers SGD test accuracy}
    \label{fig:sample-conv}
\end{figure}

We see that switching from using the average micro-batch gradient norm to using a single sample micro-batch gradient norm as a regularizer does not impact final generalization performance. However, we do see in the CIFAR100 case that while using the sample-based regularizer is faster in terms of per iteration wall-clock time, the average-based regularizer converges to an optimal performance in considerably fewer gradient update steps (Figure \ref{fig:sample-conv}).

\textcolor{red}{\subsection{Limitations of Anticorrelated Perturbed Gradient Descent}\label{appendix:pgd}}

\textcolor{red}{\citet{https://doi.org/10.48550/arxiv.2202.02831} proposes a method for improving generalization by injecting spherical Gaussian noise (with variance $\sigma^2$ as a hyperparameter) at each gradient update step that is \emph{anticorrelated} between concurrent time steps, which they term \emph{Anti-PGD}. They empirically show that on CIFAR10, training a ResNet18 in the large-batch regime with Anti-PGD and then shutting off the noise (to allow for convergence) allows them to beat small-batch SGD generalization performance.}

\textcolor{red}{However, when we extended their methodology to the CIFAR100 regime (while removing possible confounding factors such as momentum), no hyperparameter combination in the large-batch regime was able to recover SGD generalization performance. This thus represents just one example of the possible endemic issue described in \ref{sec:imps}.  Hyperparameter combinations and final test accuracy are show below.}

\begin{table}[h]
    \centering
    \textcolor{red}{\caption{ResNet18 w/Anti-PGD on CIFAR100 (SB SGD Test Accuracy = 71.21). No hyperparameter combination comes close to recovering SGD performance.}\label{tab:antipgd}
    \begin{tabular}{lll}
    \toprule
    Learning Rate ($\eta$) & $\sigma^2$ & Test Accuracy\\
    \midrule
    0.5 & 0.01 & 67.54 \\
    0.5 & 0.001 & 65.44 \\
    0.1 & 0.01 & 64.55 \\
    0.1 & 0.001 & 64.90 \\
    0.05 & 0.01 & 62.52 \\
    0.05 & 0.001 & 62.78 \\
    \bottomrule
    \end{tabular}}
\end{table}

\subsection{\change{Explicit Regularization with SOTA Optimization Tools}}\label{appendix:sota}

\change{In the present work, 
we are concerned with understanding the generalization
performance of explicit regularization mechanisms in SGD
\emph{with no other modifications.} However, in practice
many different heuristics are used to improve SGD, including
momentum \citep{sutskever2013importance}, weight decay \citep{yang2022better}, and learning rate scheduling \citep{loshchilov2016sgdr}. To verify that the behavior
seen throughout the paper holds in more standard conditions,
we return to the set-up of training a ResNet18 on CIFAR100, this
time with momentum, weight decay, and cosine annealed learning rate.
Here, we focus on the two successful regularization algorithms (i.e. 
Gradient Norm and Fisher Trace regularization). 
In Table \ref{tab:sota}, we see that the behavior shown 
in the main paper still holds: with a large gap between small-batch
and large-batch SGD, Fisher Trace regularization is able to recover 
small-batch performance, while Gradient Norm regularization is able 
to beat large-batch but not fully recover small-batch performance.
Specific values for the added hyperparameters are detailed in Appendix \ref{appendix:setup}.}

\begin{table}
    \centering
    \caption{ResNet18/CIFAR100 Test Accuracy with momentum, weight decay, and cosine annealed learning rate. The relative performance of regularized training is maintained when adding additional optimization tools.}\label{tab:sota}
    \begin{tabular}{llll}
    \toprule
    SB SGD & LB SGD & LB + GN & LB + FT\\
    \midrule
    72.59 & 67.47 & 70.58 & 72.53\\
    \bottomrule
    \end{tabular}
\end{table}

\subsection{Experimental Setup Details}\label{appendix:setup}

All experiments run for the present paper were performed using the Pytorch deep learning API, and source code can be found here: 
\url{https://github.com/ZacharyNovack/imp-regularizers-arxiv}.

\change{Values for our hyperparameters in our main experiments are detailed below:}
\begin{table}
    \centering
    \textcolor{red}{\caption{Learning rate ($\eta$) used in main experiments}\label{tab:hyperparams-main-eta}
    \change{\begin{tabular}{lllllll}
    \toprule
    Model/Dataset & SB SGD & LB SGD & LB + GN & LB + FT & LB + AJ & LB + UJ\\
    \midrule
    ResNet-18/CIFAR10 &	$\eta=0.1$	& $\eta=0.1$ & 	$\eta=0.1$ &	$\eta=0.1$ & $\eta=0.1$ & $\eta=0.1$ \\
    ResNet-18/CIFAR100 &	$\eta=0.1$	& $\eta=0.5$ & 	$\eta=0.1$ &	$\eta=0.1$ & $\eta=0.1$ & $\eta=0.1$ \\
    ResNet-18/Tiny-ImageNet &	$\eta=0.1$	& $\eta=0.5$ & 	$\eta=0.1$ &	$\eta=0.1$ & $\eta=0.5$ & $\eta=0.1$ \\
    ResNet-18/SVHN &	$\eta=0.1$	& $\eta=0.1$ & 	$\eta=0.1$ &	$\eta=0.1$ & $\eta=0.1$ & $\eta=0.1$ \\
    VGG11/CIFAR10 &	$\eta=0.15$	& $\eta=0.01$ & 	$\eta=0.01$ &	$\eta=0.01$ & $\eta=0.01$ & N/A \\
    \bottomrule
    \end{tabular}}}
\end{table}

\begin{table}
    \centering
    \textcolor{red}{\caption{Regularization strength ($\lambda$) used in main experiments}\label{tab:hyperparams-main-lambda}
    \change{\begin{tabular}{lllll}
    \toprule
    Model/Dataset & LB + GN & LB + FT & LB + AJ & LB + UJ\\
    \midrule
    ResNet-18/CIFAR10 &	$\lambda=0.01$ & $\lambda=0.01$ & $\lambda=0.001$ & $\lambda=0.001$ \\
    ResNet-18/CIFAR100 & $\lambda=0.01$ & $\lambda=0.01$ & $\lambda=5 \times 10^{-5}$ & $\lambda=0.001$ \\
    ResNet-18/Tiny-ImageNet & $\lambda=0.01$ & $\lambda=0.01$ & $\lambda=1 \times 10^{-5}$ & $\lambda=0.001$ \\
    ResNet-18/SVHN & $\lambda=0.01$ & $\lambda=0.01$ & $\lambda=0.0001$ & $\lambda=0.001$ \\
    VGG11/CIFAR10 &	$\lambda=0.5$ & $\lambda=0.5$ & $\lambda=2 \times 10^{-5}$ & N/A \\
    \bottomrule
    \end{tabular}}}
\end{table}

\begin{table}
    \centering
    \textcolor{red}{\caption{Hyperparameters for large micro-batch experiments}\label{tab:hyperparams-microbatch}
    \change{\begin{tabular}{lllll}
    \toprule
    Model / & \multirow{2}{*}{Experiment} & Microbatch & Learning  & Regularization  \\
    Dataset & & Size & Rate ($\eta$) & Strength ($\lambda$)\\
    \midrule
    ResNet-18/CIFAR100 &	LB + GN & 2560 & 0.5 & 0.0025 \\
    ResNet-18/CIFAR100 & LB + FT & 2560 & 0.1 & 0.01 \\
    ResNet-18/Tiny-ImageNet &	LB + GN & 2560 & 0.5 & 0.1 \\
    ResNet-18/Tiny-ImageNet & LB + FT & 2560 & 0.1 & 0.1 \\
    VGG11/CIFAR10 &	LB + GN & 1000 & 0.01 & 0.25 \\
    VGG11/CIFAR10 &	LB + FT & 2500 & 0.01 & 0.25 \\
    \bottomrule
    \end{tabular}}}
\end{table}

\begin{table}
    \centering
    \textcolor{red}{\caption{Hyperparameters for sample micro-batch experiments}\label{tab:hyperparams-sample-microbatch}
    \change{\begin{tabular}{lllll}
    \toprule
    Model / & \multirow{2}{*}{Experiment} & Microbatch & Learning  & Regularization  \\
    Dataset & & Size & Rate ($\eta$) & Strength ($\lambda$)\\
    \midrule
    VGG11/CIFAR10 &	SB SGD & N/A & 0.01 & N/A \\
    VGG11/CIFAR10 &	LB + FT & 50 & 0.01 & 0.25 \\
    VGG11/CIFAR10 &	LB + FT & 100 & 0.01 & 0.5 \\
    VGG11/CIFAR10 &	LB + FT & 1000 & 0.01 & 0.5 \\
    VGG11/CIFAR10 &	LB + FT & 2500 & 0.01 & 0.5 \\
    \bottomrule
    \end{tabular}}}
\end{table}

\begin{table}
    \centering
    \textcolor{red}{\caption{Hyperparameters for Grafting Experiments}\label{tab:hyperparams-grafting}
    \change{\begin{tabular}{llllll}
    \toprule
    Model/Dataset & SB SGD & LB SGD & Iterative Grafting & External Grafting & NGD \\
    \midrule
    ResNet-18/CIFAR10 & $\eta=0.1$ & $\eta=0.1$ & $\eta=0.1$ & $\eta=0.1$& $\eta=0.2626$ \\
    ResNet-18/CIFAR100 & $\eta=0.1$ & $\eta=0.5$ & $\eta=0.1$ & $\eta=0.1$& $\eta=0.3951$ \\
    VGG-16/CIFAR10 & $\eta=0.05$ & $\eta=0.1$ & $\eta=0.05$ & $\eta=0.05$& $\eta=0.2388$ \\
    VGG-16/CIFAR100 & $\eta=0.1$ & $\eta=0.1$ & $\eta=0.1$ & $\eta=0.1$& $\eta=0.4322$ \\
    \bottomrule
    \end{tabular}}}
\end{table}

\paragraph{ResNet-18} For all ResNet-18 experiments, we use the standard He initialization \citep{https://doi.org/10.48550/arxiv.1512.03385}, and the default Pytorch batch normalization initialization. Additionally, we use the standard data augmentations for CIFAR10 and CIFAR100; that is, random cropping, horizontal flipping, and whitening. \textcolor{red}{For SVHN, we performed only whitening on the dataset. For Tiny-ImageNet, no data augmentations were made aside from rescaling the input images (which are $64 \times 64$) to be $32 \times 32$. Additionally, for Tiny-ImageNet the sample regularization penalties were used rather than the normal average regularization penalties given compute constraints (see Section \ref{sec:samp} for the documented similarities between the sample and average regularizations). }

All experiments were run for 50000 update iterations. In this case, all models are trained well past the point of reaching 100\% training accuracy. No weight decay or momentum was used in \emph{any} of the experiments. We use a large-batch size of 5120 for all experiments, and thus have 40 micro-batches of 128 examples each for the regularization experiments. We calculate the penalty term at every update step, which is different from the procedure in \citet{DBLP:journals/corr/abs-2012-14193}, which recalculates the penalty term only every 10 update steps. For the external grafting experiments, we use the gradient norm data from a separate run of small-batch SGD with batch size equal to the same micro-batch size used for iterative grafting (i.e. 128). All experiments were run on a single RTX A6000 NVidia GPU.

\change{For the experiments with added optimization tools in Appendix \ref{appendix:sota}, 
we take inspiration from \citet{DBLP:journals/corr/abs-2012-14193} and \citet{DBLP:journals/corr/abs-2109-14119} 
and use momentum $= 0.9$, 
weight decay $=1\times 10^{-4}$, and a cosine annealing schedule that anneals the initial learning rate to 0 every 300 epochs. This set-up is used for all experiments in this section.}

\paragraph{VGG16} The set-up for the VGG16 experiments are identical to the ResNet-18 experiments, including the usage of batch normalization within the architecture.

\paragraph{VGG11 without batch-normalization}
For all VGG-11 experiments, we train the network with a fixed learning rate (and no momentum) until we reach 99\% train accuracy. 
Note that we do not use any form of data augmentations. 
We use a small batch size of 100 and a large batch size of 5000.

\end{document}